\documentclass{article}

\usepackage{arxiv}

\usepackage[utf8]{inputenc} 
\usepackage[T1]{fontenc}    
\usepackage{hyperref}       
\usepackage{url}            
\usepackage{booktabs}       
\usepackage{amsfonts}       
\usepackage{nicefrac}       
\usepackage{microtype}      
\usepackage{lipsum}
\usepackage{graphicx}
\usepackage{subfig}

\title{PK-GCN: Prior Knowledge Assisted Image Classification using Graph Convolution Networks}

\author{Xueli Xiao\\
  Computer Science Department\\
  Georgia State University\\
  Atlanta, GA 30303 \\
  \texttt{xxiao2@student.gsu.edu} \\
  \And
  Chunyan Ji \\
  Computer Science Department\\
  Georgia State University\\
  Atlanta, GA 30303 \\
  \texttt{cji2@student.gsu.edu} \\
  \And
  Thosini Bamunu Mudiyanselage \\
  Computer Science Department\\
  Georgia State University\\
  Atlanta, GA 30303 \\
  \texttt{tbamunumudiyanselag1@student.gsu.edu} \\
  \And
  Yi Pan \\
  Computer Science Department\\
  Georgia State University\\
  Atlanta, GA 30303 \\
  \texttt{yipan@gsu.edu} \\
}

\begin{document}
\maketitle

\begin{abstract}
Deep learning has gained great success in various classification tasks. Typically, deep learning models learn underlying features directly from data, and no underlying relationship between classes are included. Similarity between classes can influence the performance of classification. In this article, we propose a method that incorporates class similarity knowledge into convolutional neural networks models using a graph convolution layer. We evaluate our method on two benchmark image datasets: MNIST and CIFAR10, and analyze the results on different data and model sizes. Experimental results show that our model can improve classification accuracy, especially when the amount of available data is small. 
\end{abstract}


\section{Introduction}
Deep learning has been successful in various domains: computer vision \cite{He2016DeepRecognition}, speech recognition \cite{Hinton2012DeepRecognition}, natural languages processing \cite{Hochreiter1997LongMemory}, bioinformatics \cite{Zeng2019DeepEP:Proteins} and so on. And deep learning models have shown great performances on classification tasks. For example, convolutional neural networks (CNNs) are frequently used for image classification. CNN models take images as inputs, and output the probabilities of images belonging to certain classes. In multiclass image classification, Image features are extracted and learned through convolutional kernels, and eventually mapped to one class among some other classes. Typically during this process, CNNs learn from the information contained within the images only, and no inter-class relationships are incorporated in the learning. The performance of classifications can greatly be impacted by the similarity between classes. For example, a cat would have a larger chance of being misidentified as a dog, than as an airplane. By learning the similarity among classes and incorporating them in our models, we could potentially improve classification performances. 

This work attempts to incorporate class similarity information into deep learning models to improve multi-class classification performance. Specifically, we experiment on image classification using CNNs. Our method is not task specific, and has the potential of being applied to classification on other types of data besides images. There are some prior works that take class similarity into consideration. Lee et al. proposed Dropmax \cite{Lee2018DropMax:Softmax}, a method which randomly drops classes adaptively, to improve the accuracy of deep learning models. During the class dropping, classes that are more similar to the input data have larger chances of being kept. Chen et al. \cite{Chen2018GraphSpace} used word semantic similarity to calculate the underlying structures between labels, and used a graph convolutional network (GCN) augmented classifier to do classification. Their method needs external information to define a label graph and initiate node representations. Using external information can have issues. For example,  in many cases, labels’ word semantic similarity cannot reflect the real similarity of the data. And to get the embeddings for classes, we need to rely on other machine learning models. Inspired by Chen et al.’s method \cite{Chen2018GraphSpace}, we propose our method that does not rely on any external information, and can be applied to much wider ranges of classification problems. Misclassification graph derived from the validation data set can be used for class similarity, and class embeddings can be extracted from model weights.  The goal of our work is not to beat the state-of-art on the benchmark image classification dataset. Instead, we use the datasets to evaluate the effect of our method on various data and model sizes, and analyze under which scenarios does our method work the best. 
The contribution of our work is as follows.  
\begin{itemize}
    \item Our method incorporates class similarity knowledge into deep learning models to improve classification accuracy. 
    \item We use a novel way of defining class similarities using misclassification graphs on the validation dataset. In our definition, similarities have directions. 
    \item We propose a novel two-stage training model. The first stage training obtains class similarity knowledge, and the second training stage combines the original model’s output with the convoluted class similarity graph outputs, and improves classification performance. 
    \item Our method does not require extra external information. Class similarity knowledge is extracted through the dataset itself. Class and data embeddings are both obtained from trained network weights. 
    \item We analyze the effect of adding class similarity knowledge with different model and dataset sizes.
\end{itemize}

The rest of our work is organized as follows: section II is related work. Section III introduces our method: how class similarity knowledge is extracted and incorporated in the deep learning classification model. Section IV describes d experiment settings and results. Section V includes conclusion and possible future directions.

\section{Related Work}
\subsection{Classification Accuracy Improvement Methods}
There are various ways of improving classification accuracy: tune model hyperparameters\cite{Young2015OptimizingAlgorithm}\cite{Real2017Large-ScaleClassifiers}\cite{Xiao2020EfficientAlgorithm}, find more data to train the model, data augmentation\cite{Shorten2019ALearning}, add more features, transfer learning\cite{Weiss2016ALearning}, model ensemble\cite{Sagi2018EnsembleSurvey} and so on. 

Model hyperparameters play an important role in model performances. The size of the model decides how well the model can fit the data. Learning rates need to be tuned carefully to help the model converge to a better optimum \cite{Basodi2020GradientNetworks}. Convolution window decides how much local information is examined at a time. And there are many more hyperparameters that are essential to models’ performance. Deciding a good set of hyperparameters is difficult work, and there are many research works in this area \cite{Young2015OptimizingAlgorithm}\cite{Real2017Large-ScaleClassifiers}\cite{Xiao2020EfficientAlgorithm}\cite{Bergstra2012RandomOptimization}\cite{Baker2016DesigningLearning}\cite{Baldominos2018EvolutionaryRecognition}\cite{Suganuma2017AArchitectures}. 

Deep learning models also rely on the abundance of data. And data augmentation, a method which manufactures data with existing data can greatly help with model performance. Popular image augmentation methods include flipping, translating, and rotating images. And there are novel image augmentation methods such as overlaying two images \cite{Inoue2018DataClassification} and masking part of the image \cite{Takahashi2018DataCNNs}. 

Model ensemble takes advantage of multiple diverse models and combines the predictions of various models on the task. It is a very powerful technique. In Kaggle competition, it is common to see the top results utilizing ensemble \cite{Sagi2018EnsembleSurvey}. Popular model regularization technique, dropout \cite{Srivastava2014Dropout:Overfitting}, also behaves like training an ensemble of submodels. 

\subsection{Knowledge incorporation in classification tasks}
Adding information / features could also improve model performances. Sometimes the information does not come from the data itself, but other sources. There are various ways of incorporating prior knowledge into deep learning. Diligenti et al. \cite{Diligenti2018IntegratingLearning} used first order logic rules and translated them to constraints which are incorporated during the backpropagation process. Towell et al. \cite{Towell1994Knowledge-BasedNetworks} proposed a hybrid training method: first translate logic rules into a neural network and then use the neural network to train classified examples. Xu et al. \cite{Xu2018AKnowledge} derived a semantic loss function that bridges between deep learning outputs and logic constraints. Hu et al. \cite{Hu2016HarnessingRules} proposed a method that iteratively distills information in logic rules into weights of neural networks. Ding et al. \cite{Ding2018PriorApplication} integrated prior knowledge in indoor object recognition problems. Color knowledge for indoor objects and object appearance frequencies are generated in the form of vectors and used to modify deep learning model outputs. Jiang et al. \cite{Jiang2018HybridDetection}incorporated semantic correlations between objects in a scene for object detection. Stewart et al. \cite{Stewart2017Label-FreeKnowledge} performed supervised learning to detect objects using domain knowledge instead of data labels, this can be applied to problems where labels are scarce.

One kind of knowledge that can be incorporated is class similarity. In many works, label relations are incorporated in deep learning models. Word taxonomy can be used to improve image object recognition\cite{Hwang2012SemanticTaxonomies}. Associated image captions can improve entry-level labels of visual objects. Structured inferences can be made by incorporating label relations. GCN augmented neural network classifiers can be used to incorporate underlying label structures. 

\subsection{Class Similarity}
There are various ways to obtain similarity between classes. Label relations can be used to define class similarities \cite{Deng2014Large-ScaleGraphs}, where semantic similarity of labels can be computed. The problem is in many situations, label relations may not be able to fully capture the similarities between the actual data. Sometimes label relations cannot represent meaningful class similarities at all. 

Class similarity knowledge does not necessarily need to come from external sources. It may be extracted from the data itself. Arino et al.\cite{Arino2018ClassSim:Classifiers} proposed to use misclassification ratios of trained deep neural networks to get class similarities. Their proposed method uses symmetrical similarity between classes. 

\subsection{Graph Convolutional Networks}
CNNs are successful in capturing the inner patterns of Euclidean data. However, lots of data in real life scenarios exist in the form of graphs. For example, social networks are graph based: the nodes are people and the edges are the connections between them. Chemical molecules, atoms held together by chemical bonds, can naturally be modeled as graphs. Analyzing their graphical structure can determine their chemical properties. In traffic networks, points of intersections are linked together by roads, and we can predict the traffic of these intersections in future times. Images can be thought of as a special kind of graph, where adjacent pixels are connected together forming a pixel grid. When images are fed through a CNN model, the nearby pixels are being convoluted and local spatial information is retained. CNNs cannot learn from graph data with more complex relations. Similarly graph convolution can be performed on graph data, where each node can learn the weighted average of its neighbors’ information. 
A graph convolutional network (GCN) \cite{Kipf2017SEMI-SUPERVISEDNETWORKS} does the following graph convolution operation:
\begin{equation} \label{gcn}
H^{(l+1)} = \sigma(\tilde{D}^{-\frac{1}{2}}\tilde{A}\tilde{D}^{-\frac{1}{2}}H^{(l)}W^{(l)})
\end{equation}
Where $\tilde{A}$ is the adjacency matrix with self-loops, $\tilde{D}$ is the node degree matrix, $W$ is a layer-specific trainable weight matrix and sigma is an activation function. $H^{(l)}$ is the output from the $l$th layer  and $H^{(0)}$ is input. During graph convolution, each node aggregates information from its neighbors.

\section{Method}
In deep learning multiclass classification tasks, data are mapped to one of the predefined classes. The model extracts features from data and no inter-class relationship is considered. Prior works have shown that incorporating class similarity knowledge can improve classification performances \cite{Lee2018DropMax:Softmax}\cite{Arino2018ClassSim:Classifiers}. We incorporate class similarity knowledge into deep learning models using graph convolution to improve classification accuracy. 

The class similarity knowledge is extracted directly through training from data, and no additional external information is required. Information directly learned from data should represent similarity more accurately than external ones. We train the model on the training dataset, and obtain the misclassification graph on the validation dataset. The misclassification graph contains information about how often one class is misclassified as another class. If data in one class is frequently misclassified as another class, we consider that the former class is similar to the latter class. 
  
The vector representations of classes and data are extracted from learned model weights and hidden layer outputs respectively. Together with the class similarity graph, class and data representations are sent to a graph convolution layer. The graph convolution adjusts the results according to class similarity knowledge, and class scores are finally sent to softmax activation function to get the final classification results.

\subsection{Represent Class Similarities Using Misclassification Graph}
We use misclassification graphs to represent class similarities. If data in one class is often misclassified as another class, we consider the two classes have high similarity. Our misclassification graph is directed, which means similarities can be directional. Class A can be very similar to Class B, but not the other way around. This directional similarity can be observed from the misclassification information. In experiments we can observe one class being frequently misclassified as another class, but not the reverse way. 

\begin{figure}[t]
\centerline{\includegraphics[width=0.5\columnwidth]{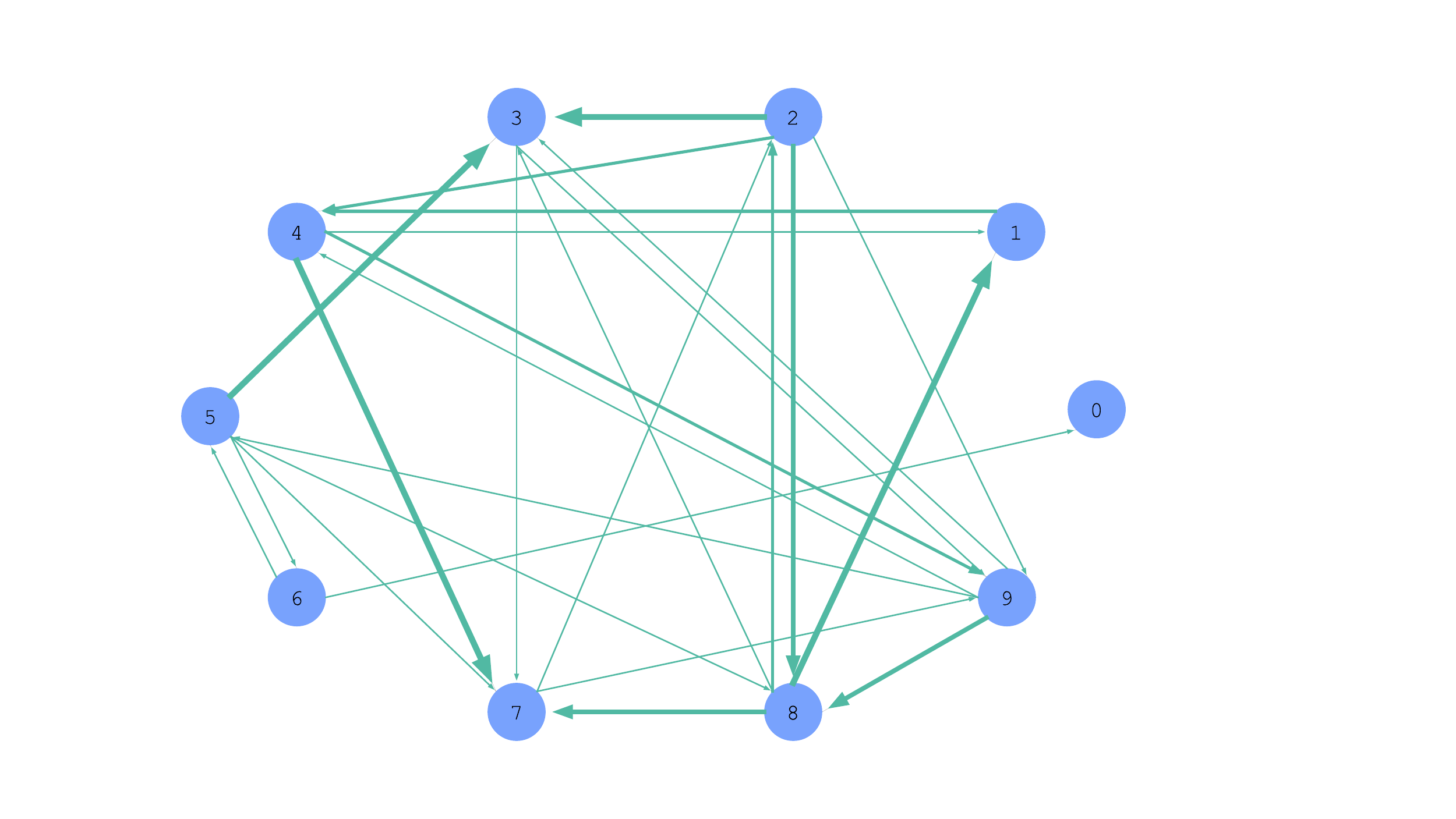}}
\caption{A Misclassification Graph example on the MNIST Dataset.}
\label{misclass}
\end{figure}

The misclassification graph is built based on a trained model’s performance on the validation dataset. Figure \ref{misclass} is an example of a misclassification graph. A CNN model is trained on a downsampled MNIST dataset. After the model is trained, we evaluate its performance on the validation dataset. Mistakes made on the validation dataset are recorded and plotted as a graph. Each node represents a class in the dataset, and edges denote how frequent data from one class are being misclassified as another class. The thicker the edge, the more misclassification between the two classes. As shown in Figure \ref{misclass}, there are 10 classes in the MNIST dataset: 0 - 10. Edges between some classes are thicker, for example: Class 8 to Class 1. This means lots of images that are actually 8’s are mistaken as 1’s. Note that edges have directions. While 8’s are easily mistaken as 1’s, barely any 1’s are misclassified as 8’s. 

\subsection{Overall Model Architecture}
\begin{figure}[t]
\centerline{\includegraphics[width=0.9\columnwidth]{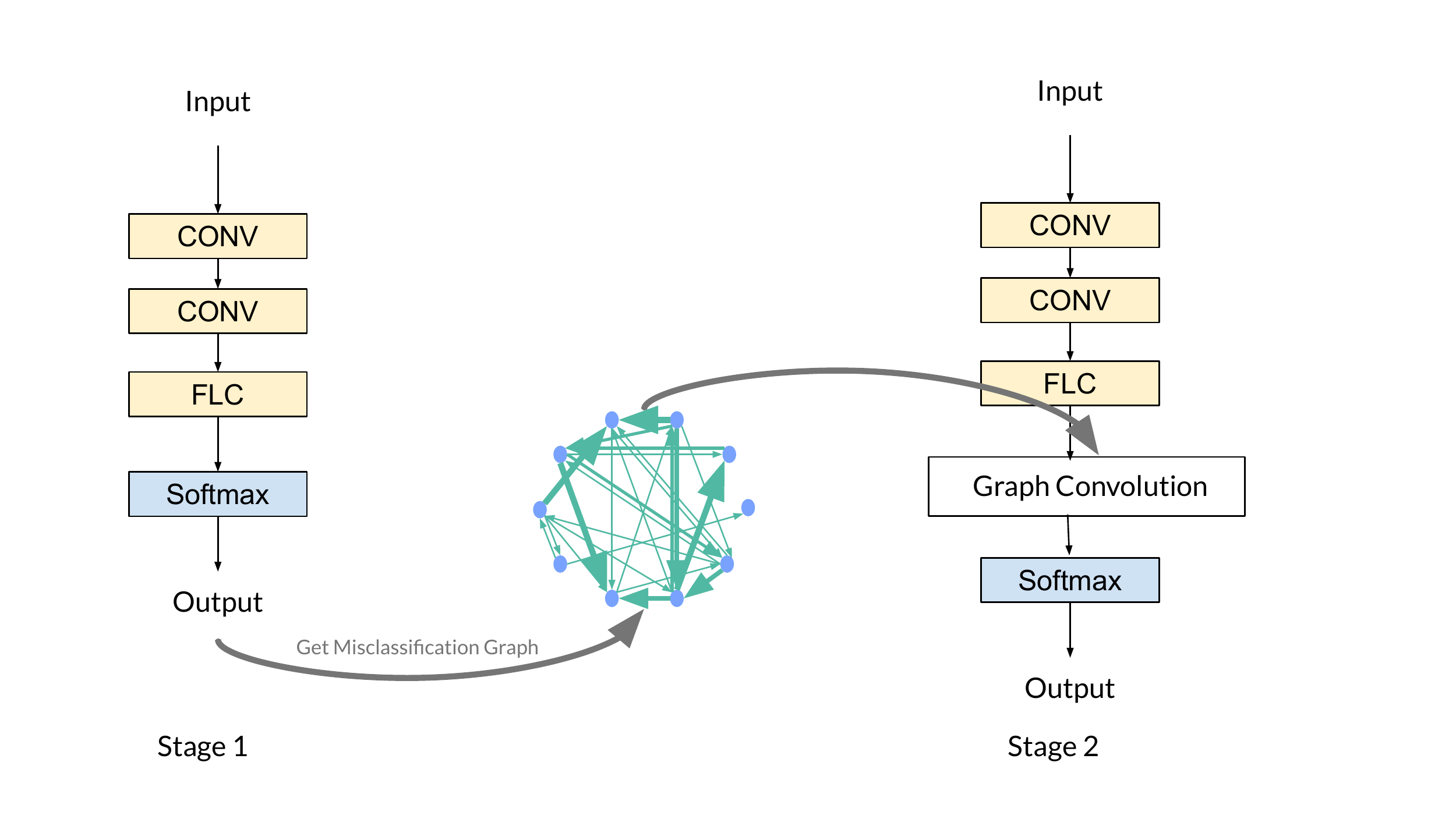}}
\caption{Overall model architecture. In Stage 1 training, mislassification graph is obtained from the the validation data. The graph represents class similarity information and is fed into the next stage. In Stage 2 training, a graph convolution layer is added to incorporate class similarity knowledge into the training.}
\label{arch}
\end{figure}

Figure \ref{arch} shows the overall architecture of our model. There are two stages of training. In Stage 1, we train an original CNN model without the graph convolution layer. In the illustration we use a CNN model with two convolutional layers and a fully connected layer as an example.  The model is trained for enough epochs such that it has learned the training data well. Then a misclassification graph is obtained by feeding the validation data through the model. Data and class embeddings are also extracted from the CNN model to produce vector representations for graph nodes. After we obtain the misclassification graph, and node representations, we can enter Stage 2 of training. In this stage, a graph convolution layer is added after the fully connected layer. The graph convolution contains the misclassification information. This layer takes in the latent data representation, and class embedding information and does convolution among the classes, and outputs new data and class embeddings with aggregated neighborhood information. The new data and class embeddings are further used to calculate class scores and finally sent to the softmax activation function to produce final results.  

\begin{figure}[t]
\centerline{\includegraphics[width=0.7\columnwidth]{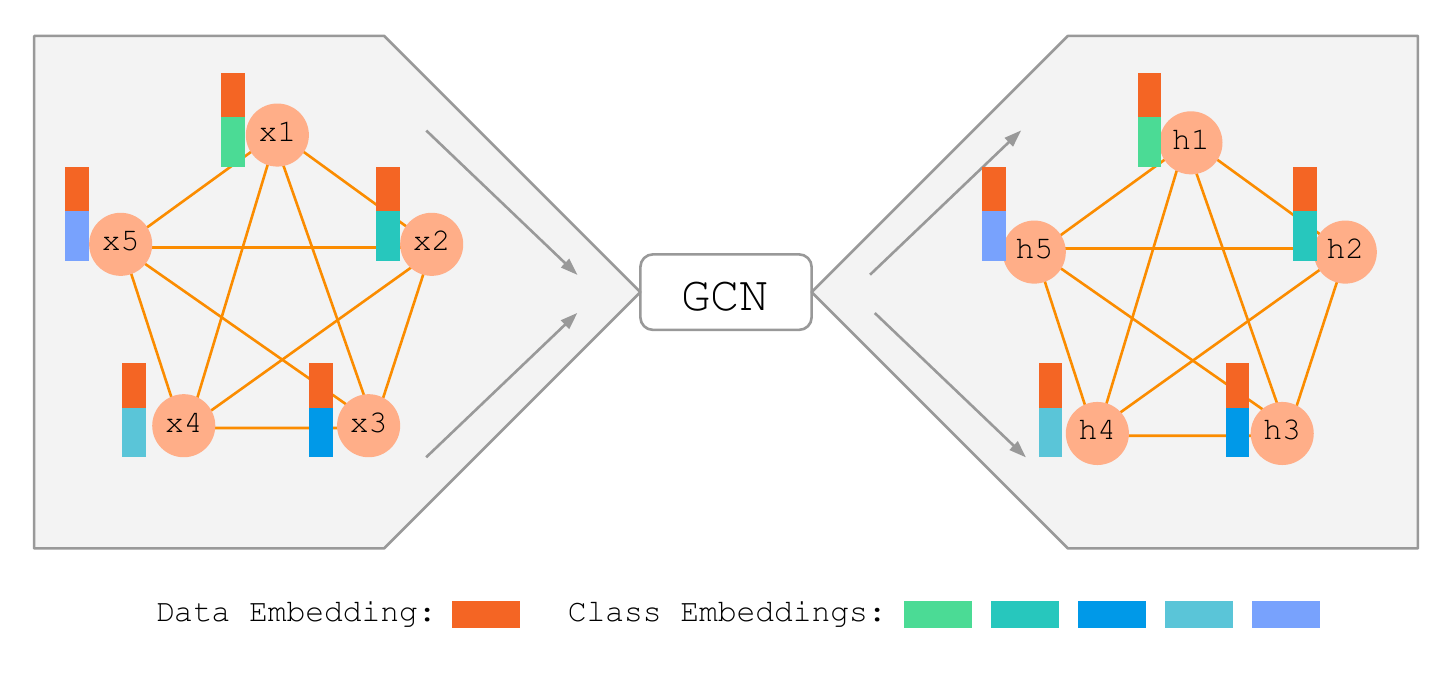}}
\caption{The graph convolutional network takes in the data embedding concatenated with class embeddings, performs neighborhood aggregation, and outputs new node representations.}
\label{gcn_fig}
\end{figure}

The graph convolution layer takes in data and class embeddings, and outputs new vectors that contain aggregated neighborhood information. Figure \ref{gcn_fig} shows a five-class graph convolution example. The nodes in the graph represent individual classes. There are five nodes in this example corresponding to five classes. The edges between nodes represent how often one class is misclassified as another class. Edges have directions. Nodes are represented using vectors of numbers. In our case, the vectors are concatenations of data embedding and class embeddings. 

\begin{figure}[t]
\centerline{\includegraphics[width=0.7\columnwidth]{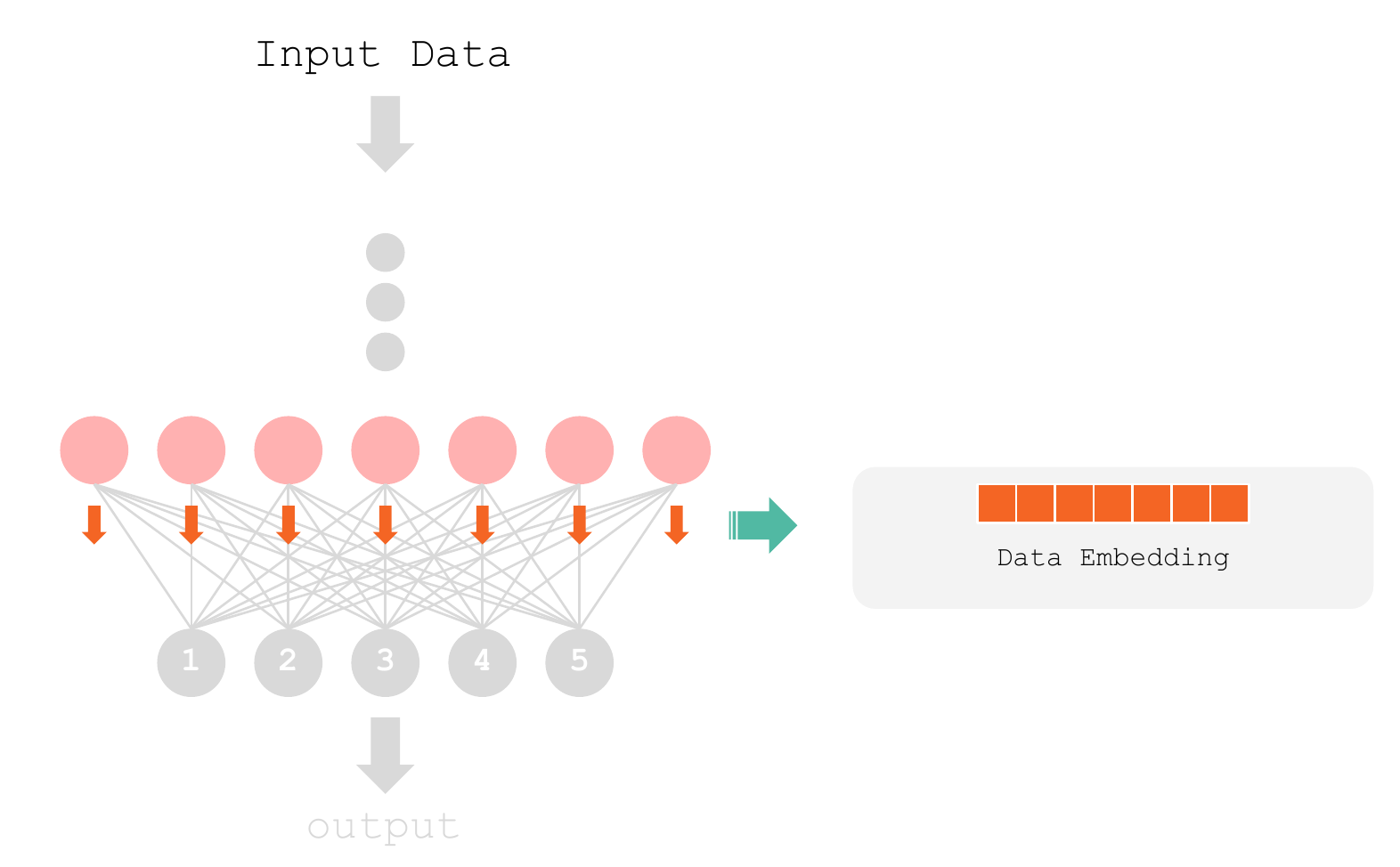}}
\caption{Obtain data embedding from the model.}
\label{data}
\end{figure}

\begin{figure}[t]
\centerline{\includegraphics[width=0.7\columnwidth]{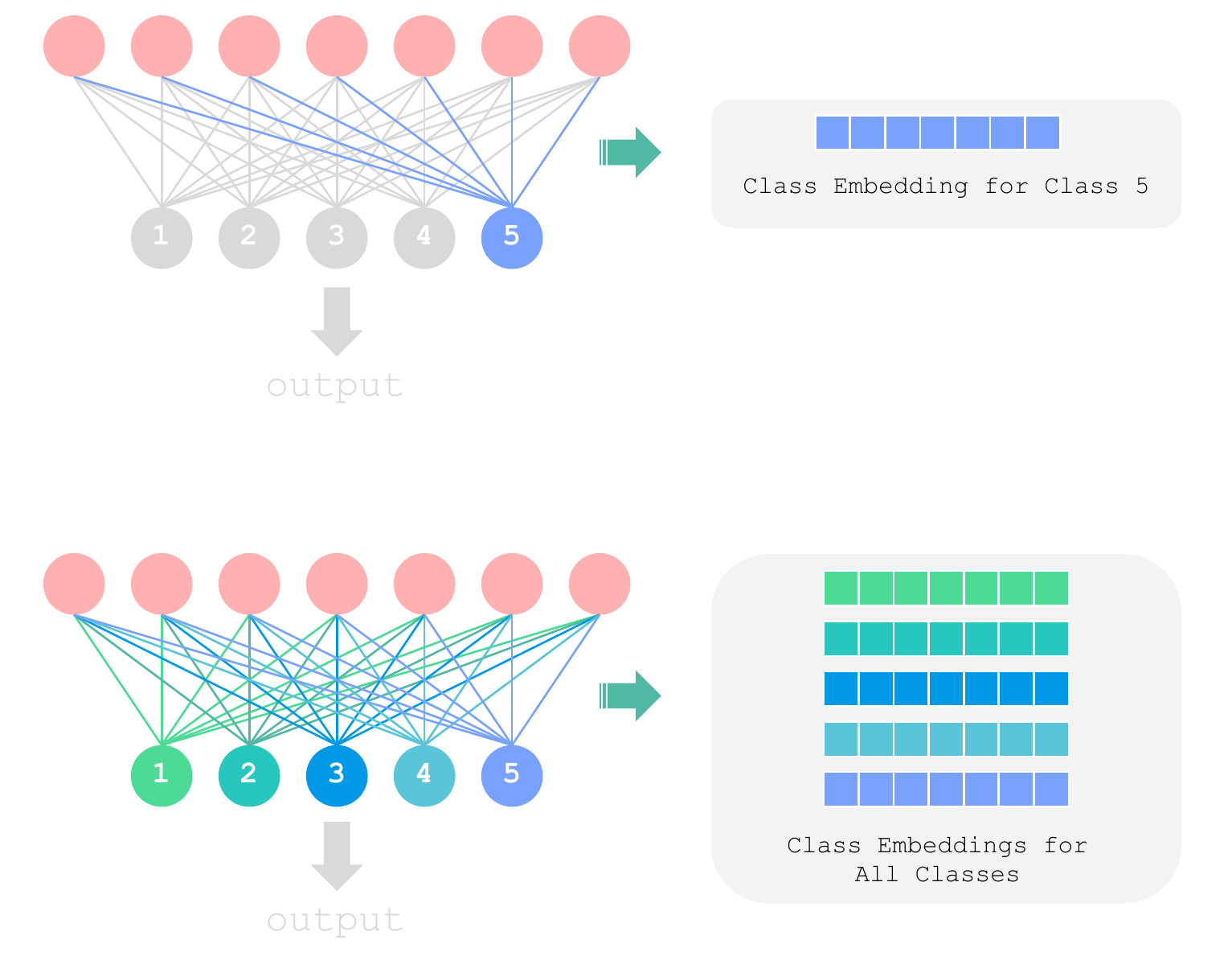}}
\caption{Obtain class embeddings for all classes from the model. }
\label{class}
\end{figure}

Both data embedding and class embedding can be obtained from the original CNN model. Data embeddings are obtained by getting the outputs from the layer right before Softmax classifier, as shown in Figure \ref{data}. Class embeddings are obtained from the weights connecting the classifier and the layer before. Figure \ref{class} shows how the class embeddings are obtained. Notice that the data and class embeddings have the same dimensions. And in the original CNN model, the inner products of data and class embeddings produce class sores. 

Data embedding and class embedding together form the graph node vector representations.

After the node vector representations pass through the graph convolution layer, the graph convoluted outputs are further transformed to get class scores before sent to the Softmax activation function. Vector representations for data and classes are extracted from the original CNN model. The data embedding for data $d$ is $\vec{d} = \{d_1,...d_k, ... d_n\}$. The class embedding for class $i$ is $\vec{c_i} = \{c_{i1},...c_{ik}, ... c_{in}\}$. In the original CNN model, the dot product of $\vec{d}$ and $\vec{c_i}$ produces the class $i$ score of $d$. The vector representation for node $i$ is the concatenation of $\vec{d}$ and $\vec{c_i}$: $\vec{x_i} =  \{d_1, ... d_n,... c_{i1},... c_{in}\}$. After graph convolution, the output for node $i$ is: $\vec{h_i} = \{h_1,... h_n,h_{n+1},... h_{2n}\}$. We need to transform the outputs from graph convolution to class scores before feeding them to softmax activation function. 

\begin{figure}[t]
\centerline{\includegraphics[width=0.5\columnwidth]{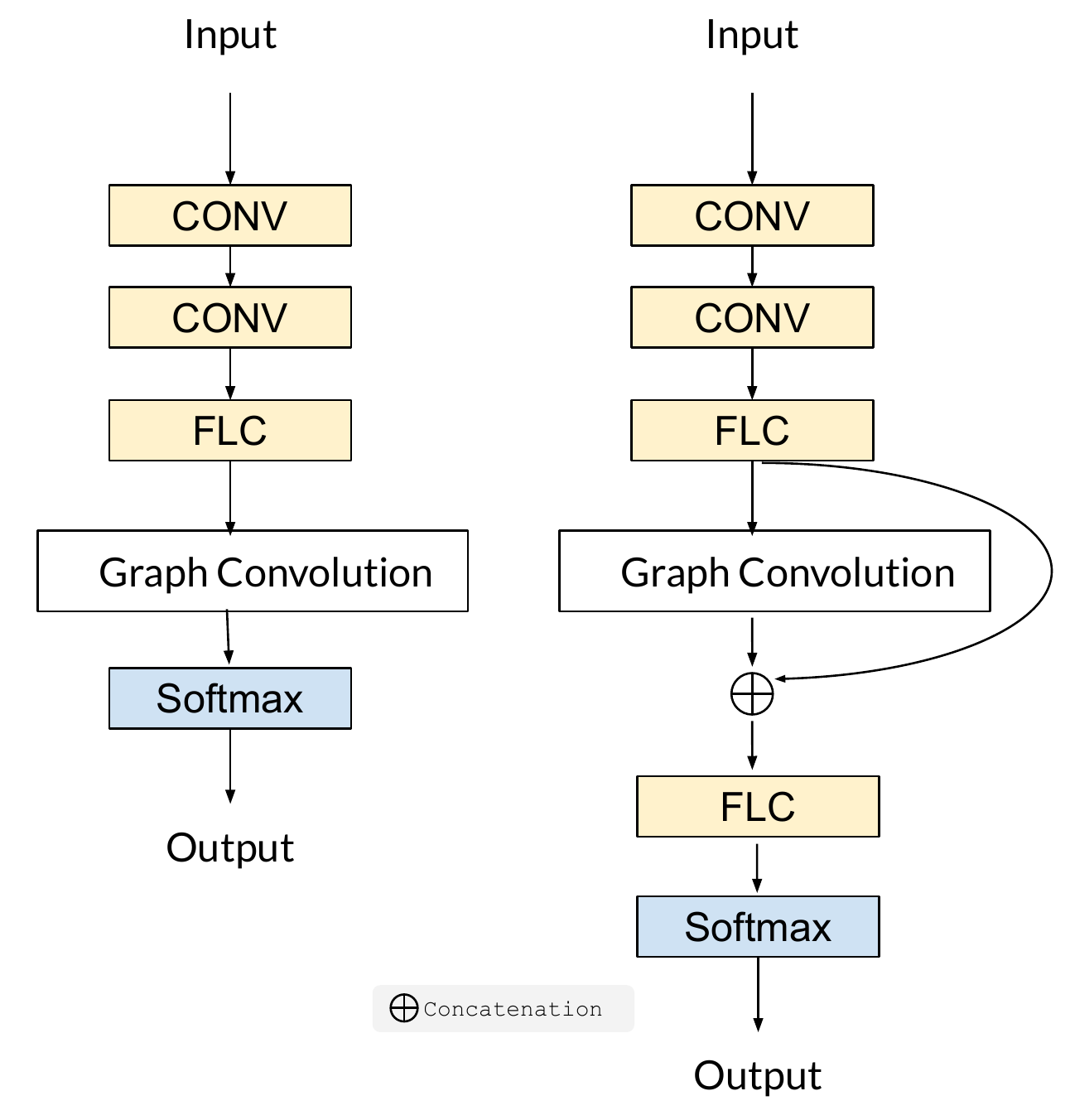}}
\caption{Two ways of incorporating the graph convolution layer.}
\label{two}
\end{figure}

We use two ways of transforming graph convoluted outputs to class scores. This will give us two variants of graph convolution assisted model: PK-GCN-1 and PK-GCN-2. Figure \ref{two} describes the difference between these two variants. In PK-GCN-1, the outputs of the graph convolution layer are directly used for producing class scores. In PK-GCN-2, a fully connected layer is added after the graph convolution layer. The fully connected layer merges the inputs and outputs of the graph convolution layer. The input to the fully connected layer is the input to the graph convolution layer concatenated with the output of the graph convolution layer. 

In PK-GCN-1, we produce class score $s_i$ for class $i$ according to the following formula:

\begin{equation} \label{gcn1}
s_i = \sum_{k=1}^{k=n}d_k h_{i(n+k)} + \sum_{k=1}^{k=n}c_{ik} h_{ik}
\end{equation}

For PK-GCN-2, we add a fully connected layer after the graph convolution layer. This layer merges the original data and class embeddings and the graph convoluted output together. Let the dimension of the output of the fully connected layer be $2l$. We use the following formula to produce class score $s_i$ for class i:

\begin{equation} \label{gcn2}
s_i = \sum_{k=1}^{k=l}q_k q_{k+l}
\end{equation}

where the output of the fully connected layer is $\vec{q}$, and its dimension is $2l$.

Our method can be summarized into the following steps:
\begin{enumerate}
    \item Train a base CNN model until convergence. This is stage 1 training. 
    \item Feed the validation dataset through the CNN model to get the misclassification graph. The graph is weighted and bidirectional. 
    \item Extract vector representation of each of the $m$ classes $\vec{c_1}$, $\vec{c_2}$, … $\vec{c_m}$ from trained model weights. (Figure 5)
    \item Obtain vector representations of data $d$ from the last hidden layer outputs. (Figure 4) 
    \item Add a graph convolution layer to the base CNN model. Node $i$ in the graph is represented by $\vec{x_i}$ which is data representation (from step 4) concatenated with class $i$ representation (from step 5). $\vec{x_i} = \{\vec{d}, \vec{c_i}\}$. 
    \item (PK-GCN-2 Only) Add a fully connected layer after the convolution layer. 
    \item (PK-GCN-1 Only) Produce class scores using equation 2.
    \item (PK-GCN-2 Only) Produce class scores using equation 3. 
    \item Continue training the model with the new layers until convergence. This is stage 2 training. 
\end{enumerate}

\section{Experiments and Results} 
We perform our experiments on MNIST dataset and CIFAR-10. In our experiment, different CNN model size and different data size is experimented to evaluate how adding class similarity knowledge helps in different circumstances. 

\subsection{Datasets}
The MNIST dataset contains 10 classes of handwritten digits: 0-9. The dataset provides a train-test split, with 60000 training and 10000 testing. The CIFAR-10 \cite{Krizhevsky2009LearningImages} dataset contains 60000 color images of size 32X32 in 10 classes: each class has 6000 images. There are 50000 images for training, and 10000 for testing.

\subsection{Baseline}
We use a two-stage training method for our models. In stage one training, a base CNN model is used to obtain the misclassification graph for class similarity. In stage two, a graph convolution layer is added to incorporate class similarity knowledge. The baselines for our method  are the base CNN models we use in stage one training. We make sure the baseline and our proposed model are trained for the same number of epochs using the same optimizer. 

\subsection{Results on Mnist Dataset}
We evaluate our method on the MNIST dataset. We experimented with two CNN base models. The first contains 1 convolution layer, the second contains two convolution layers. We also experiment with different train and validation data sizes. We make sure that the training and validation dataset have a balanced number of data from each class. We report the accuracy of the baseline and our models on the test dataset, which contains 10000 images. 

Table 1 shows the results comparison between our model and the original CNN. When using base model 1, the original CNN model was trained for 200 epochs. PK-GCN-1 and PK-GCN-2 were trained for 40 epochs in the first training stage, and 160 epochs in the second.When we use base model 2, the original CNN model was trained for 200 epochs. PK-GCN-1 and PK-GCN-2 were trained for 80 epochs in the first training stage, and 120 epochs in the second. All training uses AdaDelta optimizer with the same setup. And all models were trained for the same number of epochs for fair comparison. 

From the table, we can see that our model outperforms the base model by as much as 1.56. And generally, the improvements are bigger when the amount of available training data is smaller. 

\begin{table}[htbp]
\begin{center}
\caption{Accuracy comparison on MNIST of the original CNN model and PK-GCN models with various data sizes. ‘X’ Means no accuracy improvement is observed.}
\begin{tabular}{ccccccccc}
\toprule
   & \textbf{Data size} & \textbf{300|} & \textbf{500|} & \textbf{1000|} & \textbf{1500|} & \textbf{2000|} & \textbf{2500|} & \textbf{3000|}\\
   
   & \textbf{(Train|Validation)} & \textbf{300} & \textbf{500} & \textbf{1000} & \textbf{1500} & \textbf{2000} & \textbf{2500} & \textbf{3000}\\


\hline
 Base model 1: & Original CNN & 85.30\% & 88.78\% & 93.07\% & 94.42\% & 95.13\% & 95.78\% & 96.26\%\\
1 conv layer \\
\cline{2-9} 
 & PK-GCN-1 &85.74\% & 89.49\% & 93.97\% & 94.90\% & 95.62\% & 96.03\% & 96.32\%\\
&& (+0.44) & (+0.71) & (+0.9) & (+0.48) & (+0.49) & (+0.25) & (+0.06) \\
\cline{2-9} 
& PK-GCN-2 & 86.28\% & 89.51\% & 94.12\% & 94.94\% & 95.66\% & 95.98\% & 96.50\%\\
&& (+0.98) & (+0.73) & (+1.05) & (+0.53) & (+0.53) & (+0.2) & (+0.24)\\

 \hline
 Base model 2 & Original CNN & 89.67\% & 92.08\% & 95.43\% & 96.00\% & 97.00\% & 97.44\% & 97.70\%\\
 2 conv layers \\
\cline{2-9} 
& PK-GCN-1 & 91.16\% & 93.46\% & x & 96.30\% & 97.01\% & 97.46\% & 97.74\% \\
&& (+1.49) & (+1.38) && (+0.30) & (+0.01) & (+0.02) & (+0.04) \\
\cline{2-9} 
& PK-GCN-2 & 91.23\% & 93.19\% & x & 96.26\% & x & x & 97.80\%\\
&& (+1.56) & (+1.11) && (+0.26) &&&(+0.10)\\

\bottomrule
\end{tabular}
\label{tab1}
\end{center}
\end{table}


 
 
 


\subsection{Results on Cifar-10 Dataset}
We evaluate our method on the CIFAR-10 dataset. The base model we use is VGG-11. We experiment with various data sizes and the models are evaluated on a test dataset with 10000 images. Table 2 shows the results comparison between our model and the original CNN on CIFAR-10. The original CNN model is trained for 300 epochs. PK-GCN-1 and PK-GCN-2 were trained for 100 epochs in the first training stage, and 200 epochs in the second. All training uses AdaDelta optimizer with the same setup. 

When using VGG-11 as the base model, we can see that our model outperforms the baseline by as much as 2.55. 

\begin{table}[htbp]
\begin{center}
\caption{Test accuracy comparison on CIFAR-10 of the original CNN model and PK-GCN models with various data sizes.}
\begin{tabular}{ccccccccc}
\toprule
   & \textbf{Data size} & \textbf{300|} & \textbf{500|} & \textbf{1000|} & \textbf{1500|} & \textbf{2000|} & \textbf{2500|} & \textbf{3000|}\\
   
   & \textbf{(Train|Validation)} & \textbf{300} & \textbf{500} & \textbf{1000} & \textbf{1500} & \textbf{2000} & \textbf{2500} & \textbf{3000}\\


\hline
 Base model: & Original CNN & 33.34\% & 35.51\% & 44.33\% & 50.09\% & 52.81\% & 56.50\% & 60.56\%\\
VGG-11 \\
\cline{2-9} 
 & PK-GCN-1 & 35.37\% & 36.01\% & 46.88\% & 51.16\% & 53.69\% & 56.98\% & 61.04\%\\
&& (+2.03) & (+0.50) & (+2.55) & (+1.07) & (+0.88) & (+0.48) & (+0.48) \\
\cline{2-9} 
& PK-GCN-2 & 34.93\% & 37.45\% & 45.72\% & 51.21\% & 53.25\% & 56.54\% & 60.82\%\\
&& (+1.59) & (+1.94) & (+2.39) & (+1.12) & (+0.44) & (+0.04) & (+0.26)\\

\bottomrule
\end{tabular}
\label{tab2}
\end{center}
\end{table}

\section{Conclusion and Future Work}
\label{sec:conclusion}
In our work, we define the similarity between classes using the misclassification graph produced on the validation dataset, and use a graph convolution layer to incorporate that information into training. Experiment results on benchmark image classification datasets show that incorporating class similarity knowledge can improve multi-class classification accuracy, especially when the amount of available data is small. 

Instead of obtaining the misclassification graph from validation data, rules can be used to define class relations. The relationships between classes are fuzzy. In the future, we plan to incorporate fuzzy logic \cite{Hu2020TW-Co-MFC:Data} \cite{Mudiyanselage2019DeepDetection}\cite{Liu2005AnNetworks}and rough set theories \cite{Zhang2014ASystems}\cite{Zhang2015ASystems}\cite{Zhang2016EfficientApproximations} to our work to define class relations.  A graph attention \cite{Velickovivelickovic2017GRAPHNETWORKS} layer can also be used in place of the graph convolution layer. The advantage of graph attention is that we do not need to know the edge information in the graph. The edges are learned through training. We could potentially study the edges learned by graph attention to see if they correlate to class similarities. 


\section*{Acknowledgment}
The authors acknowledge molecular basis of disease (MBD) at Georgia State University for supporting this research, as well as the high performance computing resources at Georgia State University (https://ursa.research.gsu.edu/high-performance-computing) for providing GPU resources. This research is also supported in part by a NVIDIA Academic Hardware Grant. The authors thank the Extreme Science and Engineering Discovery Environment (XSEDE)\cite{Towns2014XSEDEengineering}, which is supported by National Science Foundation grant number ACI-1548562. Specifically, the authors used the Bridges system\cite{Nystrom:2015:BUF:2792745.2792775}, which is supported by NSF award number ACI-1445606, at the Pittsburgh Supercomputing Center (PSC). The authors also thank the National Science Foundation of China (No. 61603313) and the Fundamental Research Funds for the Central Universities (No. 2682017CX097) for supporting this work.

\bibliographystyle{unsrt}  
\bibliography{references}  


\end{document}